\documentclass{article} 
\usepackage{iclr2026_conference,times}

\usepackage{amsmath,amsfonts,bm}

\def\eqref#1{equation~\ref{#1}}

\def\1{\bm{1}}

\DeclareMathAlphabet{\mathsfit}{\encodingdefault}{\sfdefault}{m}{sl}
\SetMathAlphabet{\mathsfit}{bold}{\encodingdefault}{\sfdefault}{bx}{n}

\usepackage{hyperref}
\usepackage{url}

\usepackage{algorithm}
\usepackage{algorithmic}
\usepackage{lipsum}
\usepackage{mwe}
\usepackage{xcolor}
\usepackage{amsmath}
\usepackage{amsfonts}
\usepackage{booktabs}
\usepackage{float}
\usepackage{cuted}
\usepackage[table]{xcolor} 
\usepackage{colortbl}
\usepackage{makecell}

\usepackage{subcaption}
\usepackage{colortbl}
\usepackage{caption}
\usepackage{array}

\hypersetup{
    linkbordercolor=green,
    citebordercolor=green,
    urlbordercolor=green
}

\usepackage{soul}
\sethlcolor{red!25}
\soulregister\citep7
\soulregister\citet7   
\soulregister\ref7
\soulregister\subsection7

\title{Beyond Skeletons: Learning  Animation Directly from Driving Videos with Same2X Training Strategy}

\author{
\begin{tabular}{llll}
Yuan Zeng$^{1}$ &
Yujia Shi$^{2,3}$ &
Yuhao Yang$^{1}$ &
Dongxia Liu$^{1,3}$ \\
Zongqing Lu$^{1}$ &
Wenming Yang$^{1}$ &
Qingmin Liao$^{1}$ &
\end{tabular}
\\[0.6em]
\small
$^{1}$Tsinghua University \quad
$^{2}$Harbin Institute of Technology \quad
$^{3}$Pengcheng Laboratory
\\[0.3em]
\small \texttt{\{zengy24,yangyh23,liudx24\}@mails.tsinghua.edu.cn,}
\small \texttt{24b951075@stu.hit.edu.cn} \\
\small \texttt{\{luzq,yang.wenming\}@sz.tsinghua.edu.cn,}
\small \texttt{liaoqm@tsinghua.edu.cn}
}

\iclrfinalcopy 
\begin{document}


\maketitle


\begin{figure}[h!] 
    \centering
    \includegraphics[width=0.99\textwidth]{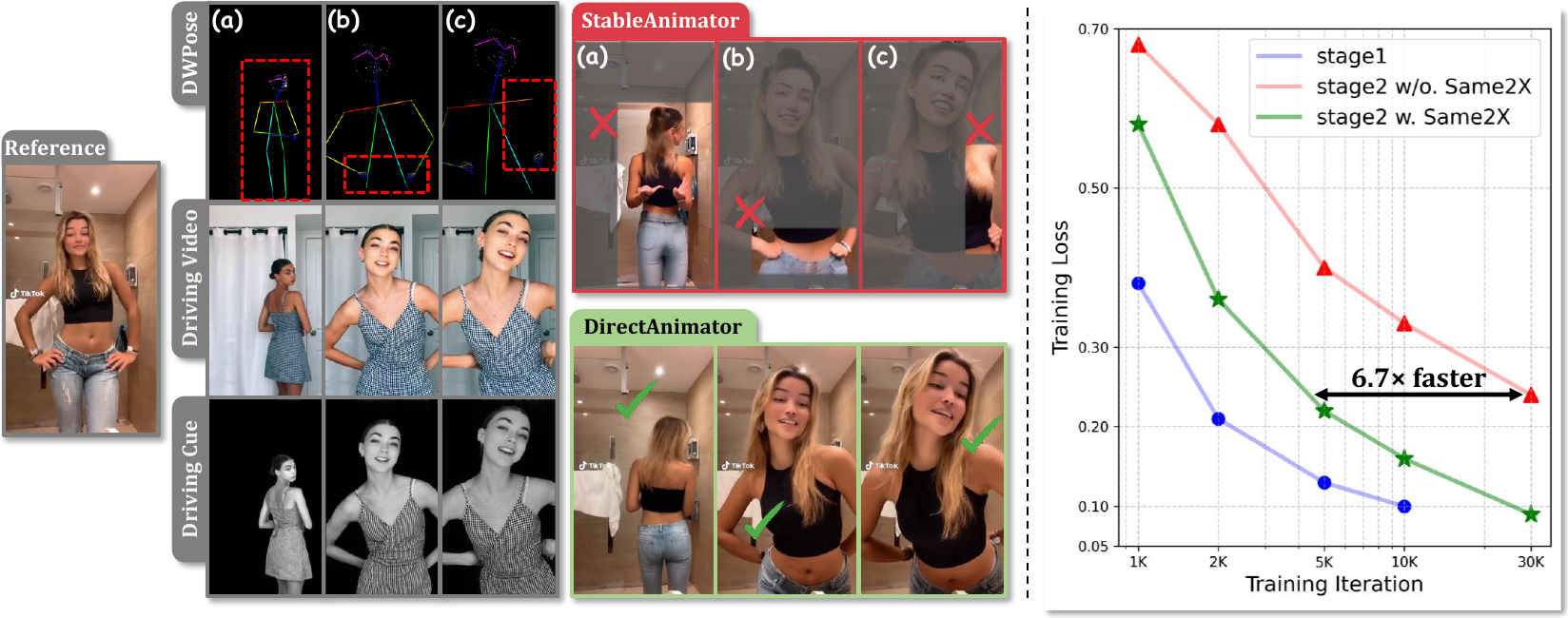}
    \caption{
    Our proposed driving cue provides a more robust representation for complex motions and self-occlusions. \textit{Left}: Errors in skeleton maps such as front-back confusion, inaccurate hand localization, and missing limbs result in noticeable artifacts in StableAnimator outputs. In contrast, DirectAnimator uses raw pixels from the driving video as driving signals, generating accurate and realistic frames. \textit{Right}: The Same2X training strategy significantly improves training efficiency in cross-ID scenarios (Stage 2), reaching the same loss level \textbf{6.7×} faster than training without it.
    }
    \label{fig:figure1}
\end{figure}

\begin{abstract}

Human image animation aims to generate a video from a static reference image, guided by pose information extracted from a driving video. Existing approaches often rely on pose estimators to extract intermediate representations, but such signals are prone to errors under occlusion or complex poses. Building on these observations, we present DirectAnimator, a framework that bypasses pose extraction and directly learns from raw driving videos. We introduce a Driving Cue Triplet consisting of pose, face, and location cues that captures motion, expression, and alignment in a semantically rich yet stable form, and we fuse them through a CueFusion DiT block for reliable control during denoising. To make learning dependable when the driving and reference identities differ, we devise a Same2X training strategy that aligns cross-ID features with those learned from same-ID data, regularizing optimization and accelerating convergence. 
Extensive experiments demonstrate that DirectAnimator attains state-of-the-art visual quality and identity preservation while remaining robust to occlusions and complex articulation, and it does so with fewer computational resources. Our project page is at 
\href{https://directanimator.github.io/}{https://directanimator.github.io/}.
\end{abstract}

\section{Introduction}
Human Image Animation (HIA) aims to generate a photorealistic video of a reference identity performing poses and expressions sourced from a driving video. This task has witnessed growing interest due to its wide range of applications in virtual avatars, video editing, and entertainment content creation. 
In recent years, numerous HIA methods have emerged, leveraging different denoising models, including SD-based \citep{sd} approaches such as AnimateAnyone \citep{Animateanyone}, SVD-based \citep{svd} methods like StableAnimator \citep{StableAnimator}, and DiT-based \citep{DiT} frameworks such as UniAnimate-DiT \citep{UniAnimateDiT}.
Most of these methods rely on intermediate representations such as skeleton maps \citep{DWPose}, DensePose \citep{Densepose}, or SMPL parameters \citep{SMPL} as driving signals to guide the animation process.
However, existing driving signals introduce several limitations. First, pose conditioning is often noisy and unreliable. Due to challenges like occlusion or complex body articulation, existing pose estimators often produce erroneous or incomplete results, leading to distorted generations. Second, pose representations often struggle to capture facial expressions accurately, as facial landmarks or sparse keypoints provide limited semantic richness, often leading to expressionless or unnatural animations.
To address these limitations, a more intuitive approach is to directly use the raw driving video as driving signal, leveraging its rich information without relying on intermediate abstractions.
This better mirrors how humans learn motion patterns, relying on holistic visual demonstrations rather than simplified pose representations.
Nevertheless, such direct conditioning is non-trivial and introduces several critical challenges.
First, in raw driving frames the cues that matter most for animation, namely body pose and facial expressions, are not explicitly encoded but are entangled with appearance details.
Unlike skeletons, where these cues are modeled by explicit keypoints and lines, the model must learn to extract and control them directly from raw pixels while keeping the reference identity intact.
Without an appropriate representation and injection mechanism, the model is prone to failing to follow the driving motion.
Second, even if motion and expression cues can be extracted in same-ID settings, training directly on cross-ID pairs further complicates optimization.
When the reference and driving identities differ, the model must simultaneously (i) follow the motion and expressions in the driving video and (ii) preserve the appearance of the reference identity.
This dual objective makes gradients noisy and slows convergence, and naive cross-ID training often leads to unstable training dynamics.

To overcome these challenges, we propose DirectAnimator, which animates reference images directly from raw driving videos. 
To address the first challenge that motion and expression cues are buried in raw pixels, DirectAnimator abstracts motion, expression, and spatial correspondence into a structured driving cue triplet and injects these cues into the denoising process via a CueFusion DiT block. 
The driving cue triplet comprises three complementary signals.
(i) Pose Cue, which captures temporally coherent pose sequences by segmenting and frequency-filtering the foreground region of the driving video;
(ii) Face Cue, which crops and centers face regions to guide expressive synthesis; and
(iii) Location Cue, consisting of grid-based softened body and face masks, provides spatial priors to facilitate alignment between the driving video and the reference image.
In the CueFusion DiT block, the reference image is encoded into vision embeddings that flow through the DiT backbone, while the three driving cues are encoded into separate control embeddings. These control embeddings modulate each DiT block via adaptive layer normalization (AdaLN) \citep{layerNorm} and gated residual connections, so that ``who the person is'' (identity) is carried by the main denoising path, and ``how the person moves and emotes'' is injected through conditional modulation. 
To further mitigate the second challenge, namely the optimization difficulty of direct video conditioning in cross-ID scenarios, we introduce the Same2X training strategy. 
The model is first pretrained on same-ID data to establish its ability to animate the reference image under the guidance of the driving cue, and then adapted to cross-ID scenarios by aligning internal representations with same-ID model via a novel Same2X alignment loss. This strategy accelerates convergence and improves generation quality in challenging cross-ID settings.

Figure~\ref{fig:figure1} demonstrates the effectiveness of our proposed driving cue and Same2X training strategy. On the left, skeleton-based methods suffer from \textbf{(a)} front–back ambiguity, \textbf{(b)} hand misplacement due to occlusion, and \textbf{(c)} missing keypoints, leading to anatomically incorrect results by StableAnimator \citep{StableAnimator}. In contrast, our Driving Cue mitigates these issues and enables more robust animation.
On the right, we show training curves of denoising loss, where our Same2X training strategy significantly accelerates convergence in the cross-ID training stage (\textit{stage 2}). It achieves the same loss level 6.7× faster than training without the strategy and reaches a final loss comparable to that of the same-ID training stage (\textit{stage 1}). Moreover, Same2X is a generalizable framework that reduces reliance on real data by leveraging pseudo data, improving training efficiency and animation quality.
Our contributions are summarized as follows:
(i)
We reformulate human image animation to use the raw driving video instead of an estimated pose proxy as the driving signal, enabling the model to learn holistic motion and expression patterns without noisy intermediate estimators.
(ii)
We instantiate this paradigm in \emph{DirectAnimator}, a framework that abstracts raw videos into a structured Driving Cue Triplet and injects these cues via a CueFusion DiT block, together with the Same2X training strategy that transfers supervision from same-ID to cross-ID settings.
(iii)
Extensive experiments and ablation studies validate the effectiveness of the DirectAnimator framework and demonstrate that it achieves state-of-the-art animation quality.

\section{Related Work}
\noindent \textbf{Driving Signals in Human Image Animation.}
Recent HIA methods explore various driving signals to improve identity preservation, motion accuracy, and generalization.
Most diffusion-based methods directly adopt 2D skeleton maps as pose conditions.
Disco \citep{Disco} is among the first to apply diffusion models to HIA, improving compositionality and generalization through disentangled control and human-centric pretraining.
AnimateAnyone \citep{Animateanyone} enhances appearance consistency and motion smoothness via ReferenceNet and pose-guided temporal modeling.
UniAnimate-DiT \citep{UniAnimateDiT} extends the UniAnimate \citep{UniAnimate} framework by replacing the SVD denoising model \citep{svd} with Wan2.1 \citep{Wan}, a DiT-based architecture for improved video synthesis.
DynamiCtrl \citep{DynamiCtrl} augments the driving signal with detailed textual prompts describing appearance, improving identity preservation via the text branch in the MM-DiT framework.
Although each method introduces unique architectural components or improvements, they all rely on 2D skeleton maps as pose conditions.
To improve the reliability of pose guidance, MimicMotion \citep{Mimicmotion} encodes keypoint confidence directly into the visual skeleton map, allowing the model to focus on reliable skeletons under occlusion and noise.
StableAnimator \citep{StableAnimator} further incorporates ArcFace-based \citep{Arcface} face embeddings into the driving signal to enhance identity preservation.
MagicAnimate \citep{Magicanimate} replaces sparse skeleton maps with DensePose maps \citep{Densepose}, enabling more robust pose conditioning for complex or rotational motions.
Champ \citep{Champ} leverages the SMPL model \citep{SMPL} along with rendered depth, normal, and semantic maps to provide geometry-aware motion guidance, enabling accurate 3D shape and pose alignment.
RealisDance \citep{RealisDance} incorporates HaMeR-based 3D hand representations \citep{hamer} to enhance realism in hand animation.
HumanDiT \citep{HumanDiT} replace skeleton maps with keypoint maps, reducing spatial overlap and ambiguity in pose conditioning.
Unlike prior methods that use intermediate representations (e.g., skeleton maps, DensePose, or SMPL)  as driving signals, DirectAnimator directly conditions on raw driving video pixels, which provides richer motion information and mitigates the accumulation of pose estimation errors.

\noindent \textbf{Facial Animation without Explicit Pose Estimation.}
There exists a line of work on facial and portrait animation that also avoids explicit pose or keypoint estimation by learning motion representations directly in latent or image space.
LIA \citep{LIA} trains a self-supervised auto-encoder and animates images via linear trajectories in the latent space, without extracting explicit structural representations.
AniTalker \citep{AniTalker} targets talking-face generation and learns an identity-disentangled motion representation together with a diffusion model for diverse facial motions.
X-Portrait \citep{Xportrait} further leverages a pre-trained reenactment network to synthesize cross-identity driving frames and uses these synthetic signals to train its motion control module, improving generalization without relying on explicit pose or keypoints.
All these methods bypass explicit pose estimation but are designed for single-face or upper-body talking-head scenarios.
In contrast, our method targets full-body human image animation, where large body articulation, extensive spatial extent, and cluttered backgrounds make direct video guidance more challenging than in single-face or talking-head scenarios.

\noindent \textbf{Representations Alignment for Training DiT.}
Representation alignment has emerged as a key technique to stabilize and accelerate the training of diffusion transformers. 
REPA \citep{REPA} first introduced an alignment mechanism that regularizes the feature space of the diffusion model to match that of a pretrained vision encoder (e.g., DINOv2 \citep{Dinov2}). This approach improves sample quality and accelerates convergence by anchoring training to semantically meaningful features. 
REPA-E \citep{Repa_e} builds on REPA by extending representation alignment to enable end-to-end training of the VAE and diffusion model. By aligning intermediate features across the two modules, REPA-E accelerates training, and enhances latent space quality.
Most recently, SRA \citep{SRA} argues that meaningful representations can emerge naturally from the generative process itself. It proposes a self-distillation approach that aligns the denoised latent features of earlier layers with those of later layers, enabling implicit representation learning without any auxiliary encoders.
Inspired by these approaches, we integrate representation alignment into the HIA training paradigm and propose the Same2X training strategy, which leverages same-ID features as internal guidance for cross-ID training stage.

\begin{figure}[h!] \centering
    \includegraphics[width=0.98\textwidth]{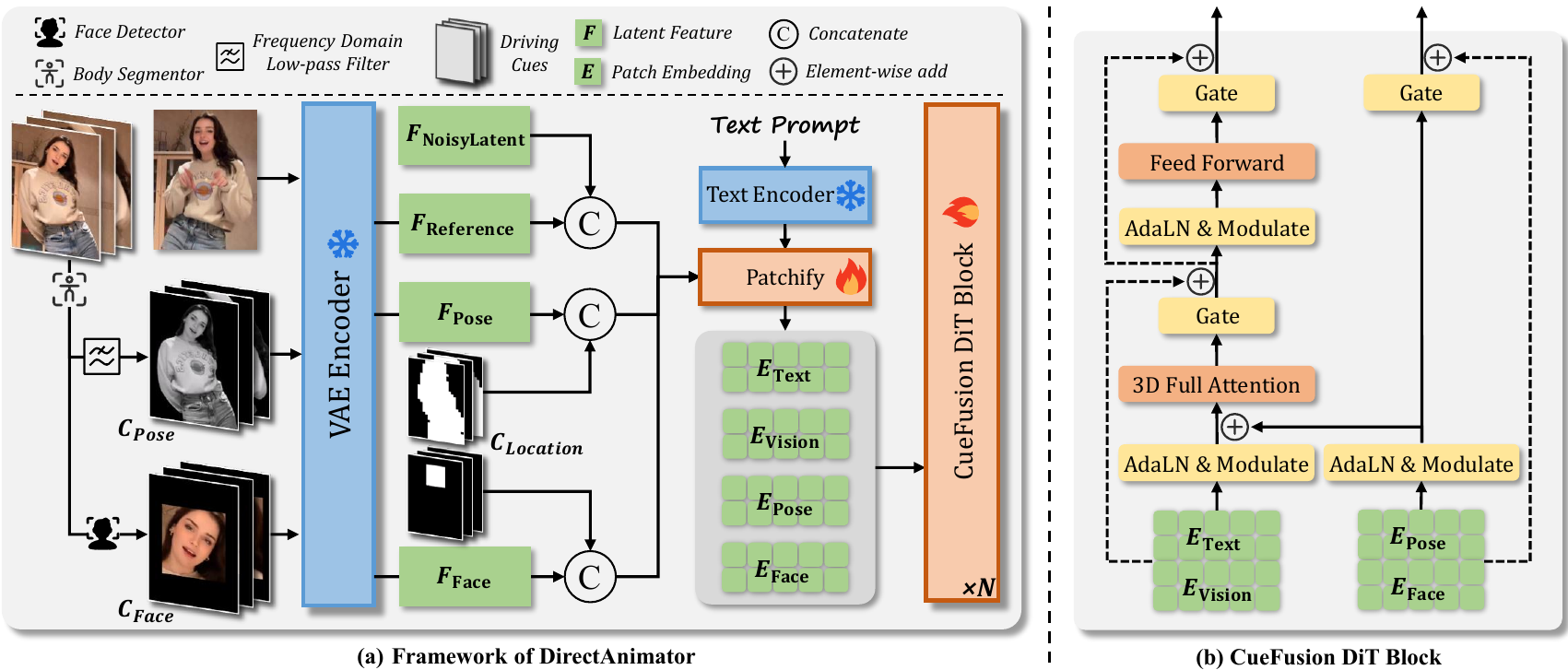}
    \caption{
    Overview of DirectAnimator.
    (a) We replace the skeleton maps with our proposed driving cue triplet: Pose Cue ($C_{\textit{Pose}}$), Face Cue ($C_{\textit{Face}}$), and Location Cue ($C_{\textit{Location}}$). A frozen VAE encoder maps the reference image, pose cue, and face cue into the latent space. Pose and face latents are each concatenated with their corresponding masks from the location cue. These features are then patchified and fed into the CF-DiT Block.
    (b) The CF-DiT Block injects pose and face cues via Adaptive LayerNorm with time-conditioned modulation, and uses gated residuals to ensure stable and controllable denoising.
    }
    \label{fig:figure2}
    \vspace{-3mm}
\end{figure}

\section{Methodology}

\subsection{Network Architecture}
\noindent \textbf{Overview.}
The overall architecture of DirectAnimator is illustrated in Figure~\ref{fig:figure2} (a), the input comprises a reference image $I$ and a driving video sequence $D_{1:N} = [D_1, ..., D_N]$. We first preprocess the driving video to extract driving cues that capture essential pose and expression information. These cues are then encoded into latent features using a 3D Variational Autoencoder (3D VAE), and subsequently transformed into a sequence of visual patch embeddings via a patchify module \citep{DiT}. The resulting embeddings are fed into the proposed CueFusion DiT Block, which performs multi-source feature fusion and denoising to generate the animated frames. 
In the following subsections, we detail the construction of the driving cues and the design of the CueFusion DiT Block.

\noindent \textbf{Driving Cue Extraction.}
Due to the inherent limitations of skeleton-based driving signal, such as instability and limited expressiveness, we propose to bypass explicit keypoint extraction and instead drive human image animation directly using raw information from the driving video. To this end, we design three complementary driving cues, namely \textit{Pose Cue}, \textit{Face Cue}, and \textit{Location Cue}, to replace traditional skeleton maps and other pose representations.
Examples of the driving cues are shown in Figure~\ref{fig:figure3}, with additional examples provided in the appendix.

\begin{figure}[h] \centering
    \includegraphics[width=0.99\textwidth]{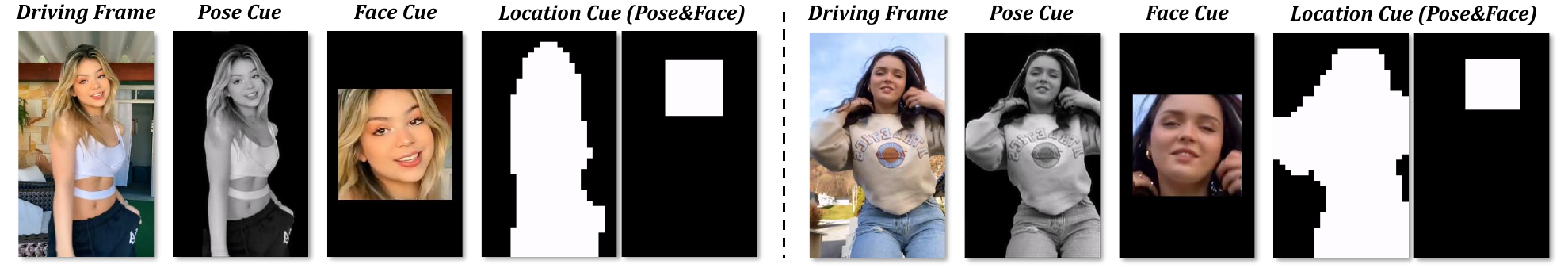}
    \caption{Examples of driving cues.}
    \label{fig:figure3}
    \vspace{-5mm}
\end{figure}

To capture the identity's pose information, we use an image segmentation model (e.g., Grounded SAM \citep{GroundedSAM}) to isolate the foreground regions from the driving video. Compared to pose estimators (e.g., OpenPose \citep{OpenPose}, DWPose \citep{DWPose}), segmentation is generally more stable under occlusion, motion blur, and other real-world conditions. Nevertheless, segmentation may still produce noisy or incomplete results. Inspired by the random dropping strategy, we discard low-quality segmentation results, forcing the model to rely on adjacent results for temporal reasoning. 
However, the segmented foreground contains rich appearance details (e.g., clothing and hair textures), such high-frequency information may distract the model from focusing on the pose information. Therefore, we apply low-pass filtering in the frequency domain to suppress irrelevant details and emphasize pose dynamics. The resulting detail-suppressed foreground sequence serves as our \textit{Pose Cue}.

Although previous methods have shown strong performance in pose transfer, their ability to capture facial expressions remains limited. We attribute this to the insufficient expressiveness of 68-point facial landmarks \citep{DWPose}, which struggle to represent complex expressions. Inspired by X-Dyna \citep{XDyna}, we directly use face regions cropped from the driving video as conditions for facial animation. Specifically, we first localize the face regions, then crop, resize, and center them to preserve maximal expressive detail. This process produces a \textit{Face Cue} that enable more expressive facial animation.

In cross-ID settings, the location and scale of the identity in the reference image may differ significantly from those in the driving video. Thus, spatial alignment is necessary before feature fusion. Prior methods typically rely on affine transformations of skeleton maps to achieve pose alignment \citep{StableAnimator, Magicanimate}. However, in DirectAnimator, since the pose cue is represented by dense foreground pixels rather than sparse skeleton maps, applying affine transformations directly may lead to distortion or structural artifacts.
Thus, we introduce \textit{Location Cue}, consisting of body and face masks. We generate body masks for the driving identity and align them to the reference identity using the spatial alignment strategy proposed in StableAnimator \citep{StableAnimator}. To mitigate the risk of identity leakage, we apply grid-based softening to the mask boundaries. Similarly, we generate and align the face mask using the same strategy.

Overall, our proposed Driving Cue, which consists of the Pose Cue (motion), Face Cue (expression), and Location Cue (alignment), not only avoids the limitations of pose estimation but also provides a more robust and semantically rich representation for animation control. Extensive experiments demonstrate its effectiveness in achieving high-quality pose transfer and expressive facial synthesis.

\noindent \textbf{CueFusion DiT Block.}
To enable the driving cue to effectively guide the denoising process of DiT \citep{DiT,Cogvideox}, we introduce the CueFusion DiT (CF-DiT) Block, whose architecture is illustrated on Figure~\ref{fig:figure2} (b). 
With the recent success of transformer diffusion models in video generation (e.g., Wan2.1 \citep{Wan}, CogVideoX \citep{Cogvideox}), a series of methods have adopted DiT as the denoising model for human video generation. These methods typically incorporate driving signals using one of the following strategies: (i) concatenating the driving features with the noisy latents as inputs to the DiT block, as in DreamActor-M1 \citep{DreamActorM1}; (ii) injecting the driving features into the vision branch via cross-attention, as in ConsisID \citep{ConsisID}; (iii) modulating the driving features using adaptive LN before injecting them into the DiT block, as in DynamiCtrl \citep{DynamiCtrl}. 

Our CF-DiT Block adopts the third strategy, which has proven to be a practical approach due to its low computational overhead and effective conditioning.
Specifically, we apply adaptive LN to inject the patch embeddings of the pose cue $e_p$ and face cue $e_p$ into the DiT block. Specifically, we first use the time embedding $e_t$ to learn modulation and gating factors for the cues via an MLP:
\begin{equation}
\alpha_p, \beta_p, \gamma_p, \alpha_f, \beta_f, \gamma_f = \textit{MLP}(\textit{SiLU}(e_t)),
\end{equation}
where $\alpha_p$, $\beta_p$, and $\gamma_p$ are the scale, shift and gating factors for modulating pose features. Similarly, $\alpha_f$, $\beta_f$, and $\gamma_f$ are factors for modulating face features.
We then use the scale ($\alpha$) and shift ($\beta$) factors to modulate the normalized cues:
\begin{equation}
\left\{
\begin{aligned}
e_p^M &= \textit{LN}(e_p) \cdot (1 + \alpha_p) + \beta_p \\
e_f^M &= \textit{LN}(e_f) \cdot (1 + \alpha_f) + \beta_f
\end{aligned} .
\right.
\end{equation}
The modulated embeddings $e_p^M$ and $e_f^M$ are added element-wise to the text and vision embeddings before entering the 3D full attention layers, preserving the original channel dimension while effectively integrating the driving cues into the denoising process.
To further ensure the stability and continuity of the driving cue throughout the denoising process, we leverage shortcut connections and gating mechanisms. The initial embeddings $e_p$ and $e_f$ are combined with the modulated ones via gated residual connections:
\begin{equation}
\left\{
\begin{aligned}
e_p^G &= e_p + \gamma_p \cdot e_p^M \\
e_f^G &= e_f + \gamma_f \cdot e_f^M
\end{aligned} .
\right.
\end{equation}
This design ensures that each DiT block receives both the raw and modulated driving cues, enhancing controllability of the denoising process.

\subsection{Same2X Training Strategy}

\begin{figure}[tb] \centering
    \includegraphics[width=0.99\textwidth]{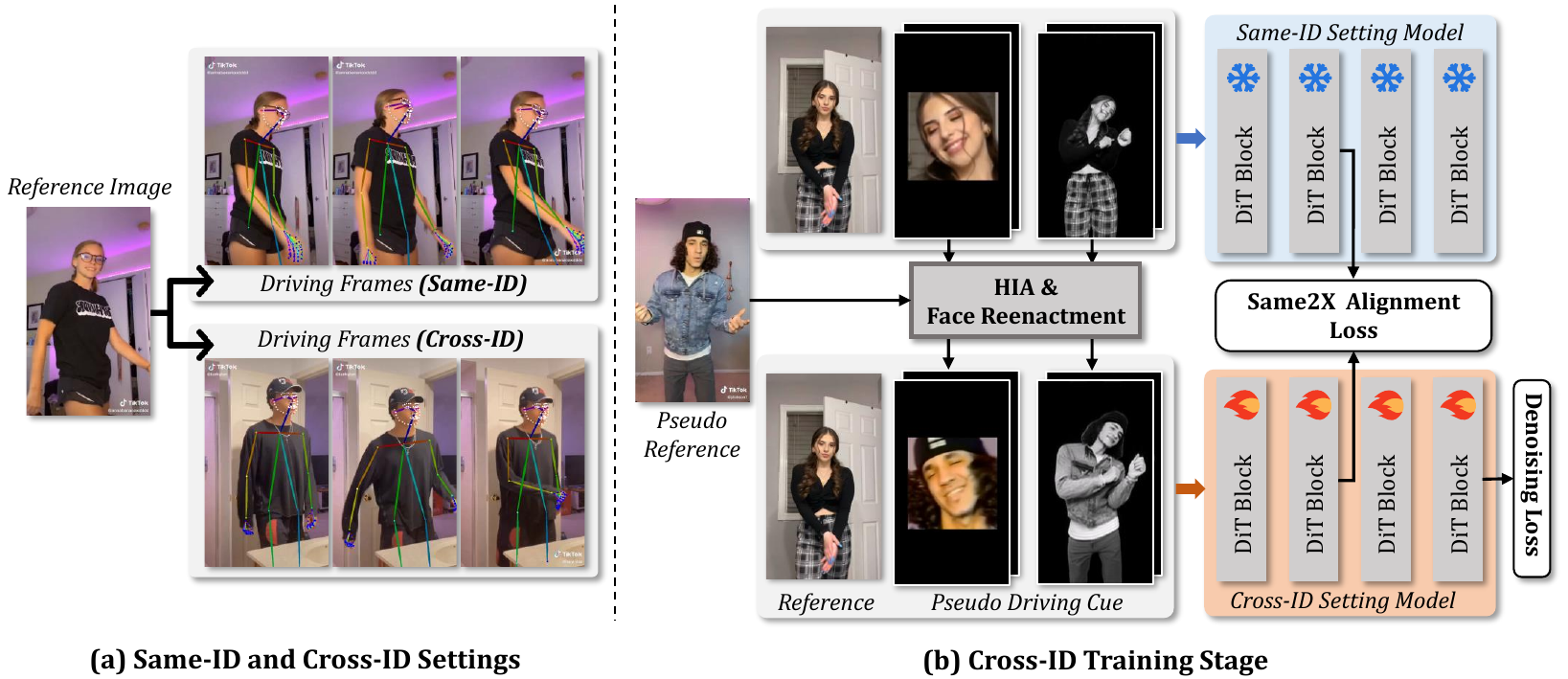}
    \caption{
    \textbf{(a)} Comparison between different settings. In the Same-ID setting, the reference image and driving video share the same identity. In the more practical Cross-ID setting, they feature different identities. 
    \textbf{(b)} Overview of the cross-ID training pipeline. First, a model is trained under the Same-ID setting. Then, in the Cross-ID training stage, a new model is trained using pseudo driving cues generated from same-ID data to simulate cross-ID conditions. The training is supervised by both a standard denoising loss and our proposed Same2X Alignment Loss. 
    This S2X loss aligns the feature embeddings of the cross-ID model with those from the pre-trained same-ID model, transferring knowledge from the simpler setting to mitigate errors in the more challenging cross-ID task.
    }
    \label{fig:figure4}
\end{figure}

The HIA task is commonly studied under two settings: same-identity (same-ID) and cross-identity (cross-ID), as shown in Figure~\ref{fig:figure4} (a). While same-ID setting is easier to train and requires less data, cross-ID setting is more practical for real-world applications.
In the cross-ID setting, where the driving and reference identities differ, most methods follow a three-step pipeline: (i) extract pose from the driving video, (ii) spatially align it with the reference identity, and (iii) use it as the driving signal.
However, this approach is suboptimal, as both pose extraction and alignment introduce unavoidable errors.


To address this, DirectAnimator bypasses pose extraction by using raw pixels from the driving video as the driving signal. While conceptually simple and flexible, directly training a model from cross-ID data is inherently challenging: the model must first understand the poses and expressions of driving identity, and then transfer them to the reference identity, making the training significantly harder.
Inspired by recent advances in training diffusion transformer (e.g., REPA \citep{REPA}, REPA-E \citep{Repa_e}, and SRA \citep{SRA}), we propose Same2X training strategy that eases training in the cross-ID setting. 
Same2X training strategy consists of a same-ID and a cross-ID training stage, with the cross-ID’s pipeline illustrated in Figure~\ref{fig:figure4} (b).
In the same-ID training stage, we train the model using pairs of reference images and driving videos from the same video clip. In the cross-ID training stage, we generate pseudo driving cues using driving videos from same-ID stage. Specifically, we use StableAnimator \citep{StableAnimator} to generate pseudo pose cues and Face-Adapter \citep{FaceAdapter} to obtain pseudo face cues. For each reference identity, we generate 0$\sim$3 pseudo driving cues to simulate cross-ID conditions. For more details on pseudo driving cue generation, please refer to the appendix.

During Cross-ID training stage, the model is supervised not only by denoising loss but also by feature alignment signals from the model trained under same-ID setting. To this end, we introduce a Same2X Alignment Loss (S2X Loss) to guide the feature dynamics:
\begin{equation}
\mathcal{L}_{\textit{S2X}}(\theta_{S}, \theta_{X}):=-\mathbb{E}_{\boldsymbol{x}, \boldsymbol{c}, \epsilon, t}\left[\frac{1}{N} \sum_{n=1}^{N} \operatorname{sim}\left(\boldsymbol{h}_{s}^{[D\_n]}, \boldsymbol{h}_{x}^{[D\_n]}\right)\right] .
\vspace{-2mm}
\end{equation}
Here, $\theta_S$ and $\theta_X$ denote the model trained under same-ID and cross-ID settings, respectively. 
$\boldsymbol{h}_s^{[D\_n]}$ and $\boldsymbol{h}_x^{[D\_n]}$ represent the patch embeddings at the $D^{\text{th}}$ DiT block for the same-ID and cross-ID models, and $n$ indexes the patch tokens. 
The function $\operatorname{sim}(\cdot, \cdot)$ measures cosine similarity.
We combine the S2X Loss and denoising loss to train DirectAnimator, with the overall loss function in the cross-ID training stage formulated as: $\mathcal{L} := \mathcal{L}_{\text{Denoising}} + \lambda \mathcal{L}_{\text{S2X}},$
where $\lambda$ is a factor controlling the contribution of the S2X loss.
As demonstrated in Figure~\ref{fig:figure1}, the Same2X training strategy significantly accelerates convergence in the cross-ID setting by alleviating the learning difficulty. 
To our knowledge, this is the first approach to leverage feature alignment across different settings for training HIA models. 

\begin{table*}[t] 
    \centering
    \newcommand{\Frst}[1]{\textbf{#1}}
    \newcommand{\Scnd}[1]{\underline{#1}}
    \caption{Quantitative comparison on the TikTok and Unseen datasets. \textbf{Bold} text indicates the best result, while \underline{underlined} text indicates the second-best. ↑ denotes that higher values are better. FIS and FTS stand for Face Identity Similarity and Face Temporal Similarity, respectively. In the table, $a/b$ denotes results on TikTok and Unseen, respectively.}
    \label{tab:table1}
    \renewcommand{\arraystretch}{1.2} 
\resizebox{1\textwidth}{!}{
\scriptsize
\LARGE
\setlength{\tabcolsep}{4pt}
\begin{tabular}{c||*{6}{c}|*{2}{c}|*{1}{c}}
\toprule
\bfseries Method                                  & \bfseries FID↓            & \bfseries SSIM↑            & \bfseries PSNR↑             & \bfseries LPIPS↓           & \bfseries L1(E-04)↓     & \bfseries FIS↑              & \bfseries FTS↑              & \bfseries FVD↓              & \bfseries Training Steps    \\  
\midrule
MagicAnimate        & 32.09 / 42.72   & 0.714 / 0.435    & 29.16 / 23.17     & 0.239 / 0.453    & 3.13 / 6.43   & 0.545 / 0.532     & 0.629 / 0.603     & 179.07 / 833.10   & -         \\ 
AnimateAnyone      & - / 36.49       & 0.718 / 0.566    & 29.56 / 24.68     & 0.285 / 0.332    & - / 4.45      & 0.516 / 0.527     & 0.642 / 0.577     & 171.90 / 785.33   & 4*A100×40K       \\
Champ                      & - / 34.62       & \Scnd{0.802} / 0.539    & 29.91 / 24.76     & 0.234 / 0.369    & 2.94 / 5.23   & 0.549 / 0.526     & 0.671 / 0.589     & 160.82 / 639.41   & 8*A100×80K   \\ 
\midrule
MimicMotion          & \Scnd{28.03} / 37.67   & 0.601 / 0.578    & - / 25.53         & 0.416 / 0.318    & 5.85 / 4.91   & 0.621 / 0.563     & 0.669 / 0.621     & 326.57 / 435.20   & 8*A100×6M     \\
StableAnimator    & - / 31.89       & 0.801 / 0.603    & \Frst{30.81} / 27.11     & 0.232 / 0.273    & 2.87 / 4.39   & \Scnd{0.662} / \Scnd{0.653}     & \Frst{0.732} / \Scnd{0.704}     & \Frst{140.62} / 365.52   & 4*A100×33M    \\ 
\midrule
UniAnimate-DiT     & - / \Scnd{29.92}       & 0.787 / \Scnd{0.649}    & 29.76 / \Scnd{27.89}     & 0.226 / \Scnd{0.261}    & 3.16 / \Scnd{3.37}   & 0.643 / 0.647     & 0.695 / 0.684     & 306.17 / \Scnd{289.45}   & -           \\
DynamiCtrl            & - / 36.80       & 0.766 / 0.633    & \Scnd{30.22} / 27.43     & \Frst{0.166} / 0.269    & \Scnd{2.34} / 3.52   & 0.632 / 0.619     & 0.725 / 0.702     & 152.31 / 339.80   & 8*H20×50K       \\
\midrule
\rowcolor{gray!15}
DirectAnimator                   & \Frst{25.87} / \Frst{27.62}   & \Frst{0.806} / \Frst{0.708}    & 30.12 / \Frst{29.41}     & \Scnd{0.215} / \Frst{0.249}    & \Frst{2.12} / \Frst{3.22}   & \Frst{0.682} / \Frst{0.661}     & \Scnd{0.730} / \Frst{0.723}     & \Scnd{142.60} / \Frst{276.34}   & 4*H20×40K         \\ 
\bottomrule
\end{tabular}
}

\end{table*}

\begin{figure*}[tb] \centering
    \includegraphics[width=0.95\textwidth]{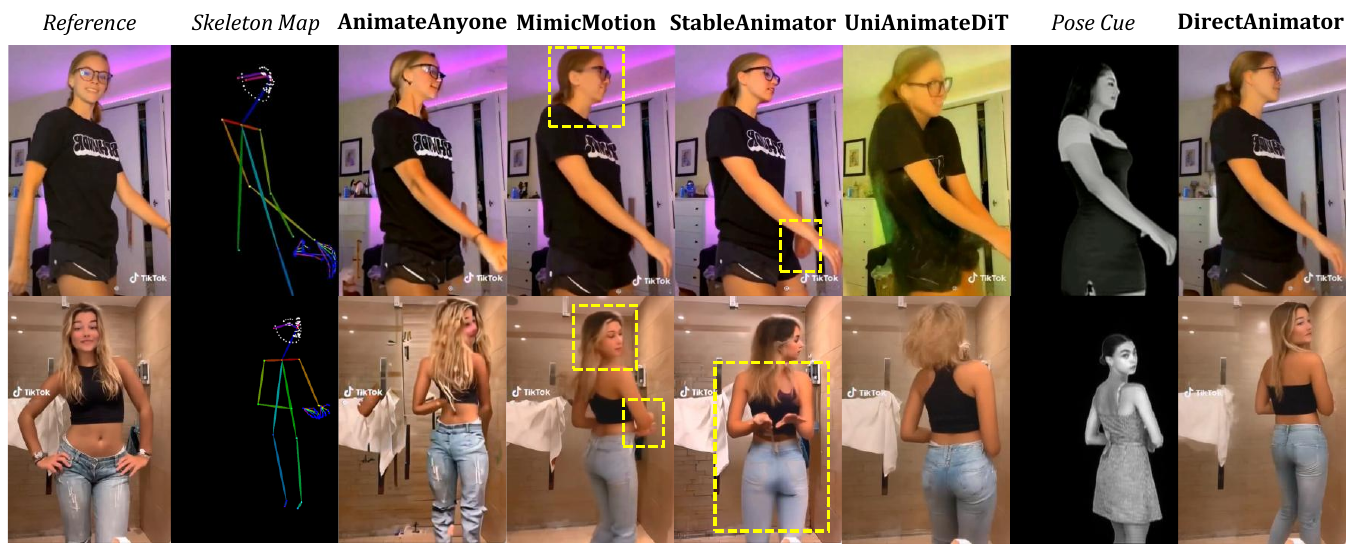}
    \caption{Qualitative comparisons between DirectAnimator and baselines on the TikTok (Row 1) and Unseen (Row 2) datasets, with visual artifacts highlighted in yellow dashed boxes for clearer comparison. User IDs are obscured for privacy protection.}
    \label{fig:figure6}
    \vspace{-5mm}
\end{figure*}

\begin{figure*}[tb] \centering
    \includegraphics[width=0.95\textwidth,]{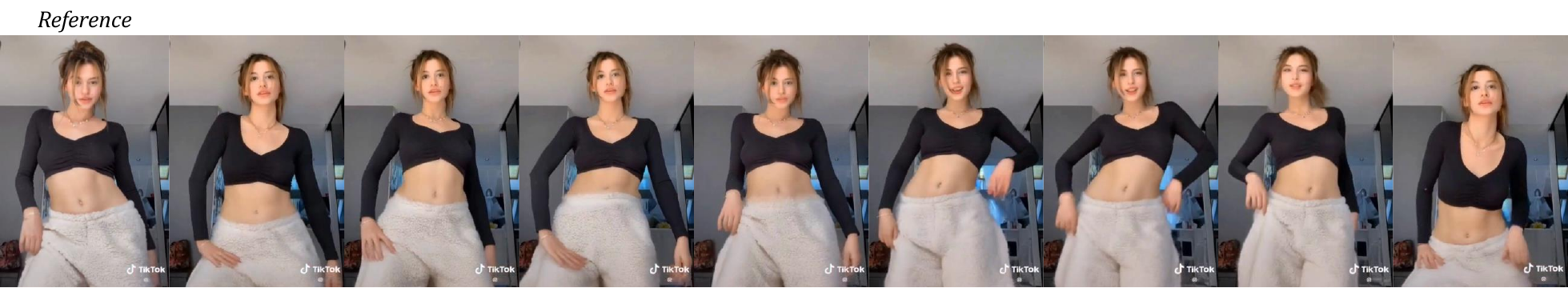}
    \caption{Sample results of DirectAnimator. User IDs are manually obscured for privacy protection.}
    \label{fig:figure7}
\end{figure*}

\section{Experiments}
\subsection{Experiments Setting}
\noindent \textbf{Implementation Details and Dataset.}
We construct our model based on CogVideoX1.5 \citep{Cogvideox}. All parameters of the DiT block are updated, while the Text Encoder and VAE Encoder are kept frozen. To accommodate video predictions of varying sizes and resolutions, we utilize a bucket sampler during training. 
The model is first trained for 10K iterations on same-ID setting with a learning rate of 2e-5, followed by 30K iterations on cross-ID setting using the same learning rate. All experiments are conducted on 4 H20 GPUs. All text prompts were generated using Qwen2-VL \citep{Qwen2} model. 
Previous works have generally not released training data due to privacy policies. Therefore, we collected 4,000 video clips (5$\sim$20 seconds) from the internet to train DirectAnimator. In addition, the training data includes video clips from the TikTok training set (1$\sim$334). All of these data are used in the same-ID training stage. For the cross-ID training stage, we synthesized a large number of pseudo data. After a rigorous filtering process, we retained 3,300 high-quality [\textit{reference, pseudo driving cue}] pairs for training.
To thoroughly evaluate the generation performance of DirectAnimator, we use video clips from the TikTok test set (335$\sim$340), along with 50 additional video clips collected from the internet (referred to as ``Unseen”), as the test set. 
To ensure fair comparison, our method is conditioned on the pseudo driving cues generated by StableAnimator, rather than using driving cues that share the same identity as the reference image.

\noindent \textbf{Baselines and Metrics.}
We conduct broad comparisons to validate our method’s superiority. Specifically, we compare against SD-based methods including AnimateAnyone \citep{Animateanyone}, MagicAnimate \citep{Magicanimate}, and Champ \citep{Champ}, SVD-based methods such as MimicMotion \citep{Mimicmotion} and StableAnimator \citep{StableAnimator}, and DiT-based methods including DynamiCtrl \citep{DynamiCtrl} and UniAnimate-DiT \citep{UniAnimateDiT}. 
We evaluate image quality using FID \citep{FID}, SSIM \citep{SSIM}, LPIPS \citep{LPIPS}, PSNR \citep{PSNR}, and L1 error. For video quality, we adopt FVD \citep{FVD}. 
We use four metrics to evaluate identity, temporal consistency, and motion transfer: Face Identity Similarity (FIS) for identity preservation, Face Temporal Similarity (FTS) for temporal consistency, Pose Landmark Consistency (PLC) for body pose following, and Facial Landmark Consistency (FLC) for facial expression and mouth-shape transfer. More detailed definitions and implementation details are provided in Appendix~\ref{pose_metric}.
Since DynamiCtrl generates only 2-second clips and is highly sensitive to text prompts, we include it in quantitative comparisons only.

\subsection{Comparison with SOTA methods}

\noindent \textbf{Quantitative Comparison.}
Table~\ref{tab:table1} presents the quantitative comparison results. As shown, DirectAnimator achieves state-of-the-art animation performance while using fewer computational resources. On the TikTok test set, our method obtains the best results in FID, SSIM, L1, and FIS, and also performs competitively in LPIPS, FVD, and FTS, indicating high visual quality and strong identity preservation. On the more challenging Unseen test set, DirectAnimator consistently outperforms all baselines across all metrics. This dataset, featuring fast-paced and complex dance motions, often leads to inaccurate pose estimations, and DirectAnimator maintains robust performance by bypassing pose extraction.
We further evaluate pose and expression transfer using PLC and FLC. As reported in Table~\ref{tab:pose_face_metrics}, DirectAnimator achieves the lowest landmark errors on both test set, confirming more accurate body pose following and more faithful facial expression transfer than existing methods. 
Finally, on a single H20 GPU at $512^2$ resolution with 49-frame clips, Table~\ref{tab:runtime} reports an end-to-end runtime comparison with key baselines to illustrate the inference cost of our design.

\noindent \textbf{Qualitative Comparison.}
Figure~\ref{fig:figure6} presents a qualitative comparison across different methods.
In Row 1, the skeleton map produces an inaccurate estimation of the driving frame: the subject’s left hand and arm should be occluded by the body and therefore not visible in the skeleton. While AnimateAnyone and MimicMotion successfully generate plausible body structures, StableAnimator mistakenly renders an extra left hand. In contrast, DirectAnimator not only produces anatomically correct results but also better preserves the subject's identity.
In Row 2, the skeleton map introduces front-back ambiguity, leading most methods to generate incorrect body orientations. Although UniAnimate-DiT generates a correct pose, it fails to model the face direction. In contrast, DirectAnimator, guided by the pose cue, produces accurate results.
Figure~\ref{fig:figure7} shows multi-frame animation results generated by DirectAnimator, which demonstrate strong identity preservation and background consistency across frames.
More results are available in the appendix.

\subsection{Ablation Study}
All ablation studies were performed on a training data subset, with both same-ID and cross-ID training stage using 500 videos each. The TikTok test set was used for evaluation.


\noindent \textbf{Driving Cues.}
To improve animation stability and reduce sensitivity to pose estimation errors, we design a Driving Cue Triplet consisting of Pose Cue (motion), Face Cue (expression), and Location Cue (alignment).
We conduct a detailed ablation study to evaluate the impact of the driving cue design on overall animation quality.
As shown in Table~\ref{tab:table2}, removing the Face Cue (\textit{S1}) causes a substantial drop in facial quality: FIS decreases from $0.638$ to $0.418$ and FVD increases from $180.52$ to $245.48$, showing that a dedicated face branch is crucial for identity and expression preservation.
Using Face Cue without enlarging and centering the detected face (\textit{S2}) recovers most of the gain, but still lags behind our full design in FIS and FVD, indicating that the enlarged and centered Face Cue yields more consistent facial details.
For the body motion signal, replacing our filtered Pose Cue with raw RGB foreground (\textit{S3}) degrades all metrics, since irrelevant appearance details are introduced into the driving signal.
A gray-scale foreground without low-pass filtering (\textit{S4}) performs better than raw RGB but remains slightly worse than our full Pose Cue, confirming that frequency-domain low-pass filtering further stabilizes pose modeling by suppressing high-frequency textures.
Substituting the entire triplet with skeleton maps (\textit{S5}) yields only moderate results and is consistently inferior to our method in FID, FIS, and FVD, which highlights the advantage of learning from carefully processed pixel cues rather than explicit pose estimators.
For spatial alignment, removing the Location Cue (\textit{S6}) leads to weaker performance, especially in FIS and FVD.
Using only a hard foreground mask as the Location Cue (\textit{S7}) partially improves over the no-Location baseline yet still falls behind our soft, grid-based Location Cue, suggesting that the softened alignment reduces artifacts and identity leakage.
These ablations demonstrate that each component of the Driving Cue Triplet contributes meaningfully to robust and high-quality HIA, and that the specific design choices are important rather than interchangeable.


\begin{table}[tb]
    \centering
    \begin{minipage}[t]{0.44\textwidth}
        \centering
        \captionof{table}{
        Pose and facial landmark consistency on TikTok / Unseen.
        }
        \label{tab:pose_face_metrics}
        \resizebox{0.9\textwidth}{!}{
        \large
        \begin{tabular}{c||*{2}{c}}
            \toprule
            \bfseries Method 
                                      & \bfseries PLC↓ & \bfseries FLC↓ \\
            \midrule
            MagicAnimate              & 0.094 / 0.128 & 0.058 / 0.079 \\
            AnimateAnyone             & 0.083 / 0.115 & 0.049 / 0.068 \\
            Champ                     & 0.098 / 0.134 & 0.054 / 0.082 \\ \midrule
            MimicMotion               & 0.081 / 0.102 & 0.045 / 0.064 \\
            StableAnimator            & 0.075 / 0.097 & 0.042 / 0.060 \\ \midrule
            UniAnimate-DiT            & 0.082 / 0.102 & 0.058 / 0.074 \\
            DynamiCtrl                & 0.075 / 0.116 & 0.042 / 0.069 \\
            \rowcolor{gray!15}
            DirectAnimator (Ours)     & \textbf{0.071} / \textbf{0.092} 
                                      & \textbf{0.037} / \textbf{0.057} \\
            \bottomrule
        \end{tabular}
        }
    \end{minipage}
    \hfill
    \begin{minipage}[t]{0.55\textwidth}
        \centering
        \captionof{table}{Ablation study for driving cues and Same2X training strategy.}
        \label{tab:table2}
        \renewcommand{\arraystretch}{1} 
        \setlength{\tabcolsep}{3pt}
        \resizebox{0.98\textwidth}{!}{
        \large
        \begin{tabular}{c l | *{5}{c}}
            \toprule
            \bfseries ID & \bfseries Settings                            & \bfseries FID↓  & \bfseries SSIM↑ & \bfseries PSNR↑ & \bfseries FIS↑ &  \bfseries FVD↓  \\
            \midrule
            S1 & \textit{w/o} Face Cue                         & 28.21 & 0.729 & 29.13 & 0.418 & 245.48 \\
            S2 & \textit{w} Face Cue \textit{(no enlarge\&center)}      & 28.02 & 0.742 & 29.45 & 0.610 & 185.93 \\
            S3 & \textit{w} Pose Cue \textit{(rgb)}            & 29.32 & 0.715 & 29.13 & 0.591 & 191.92 \\ 
            S4 & \textit{w} Pose Cue \textit{(gray, no low-pass)}       & 28.10 & 0.744 & 29.40 & 0.625 & 184.07 \\
            S5 & \textit{w} Skeleton Map                       & 29.74 & 0.710 & 29.01 & 0.578 & 216.38 \\
            S6 & \textit{w/o} Location Cue                     & 30.59 & 0.682 & 29.17 & 0.529 & 203.72 \\
            S7 & \textit{w} Location Cue \textit{(hard mask only)}      & 29.13 & 0.718 & 29.25 & 0.582 & 228.54 \\
            S8 & \textit{w/o} Same2X                           & 32.21 & 0.691 & 28.67 & 0.530 & 290.43 \\
            \rowcolor{gray!15}
               & DirectAnimator (Ours)                         & 27.61 & 0.752 & 29.53 & 0.638 & 180.52 \\
            \bottomrule
        \end{tabular}
        }
    \end{minipage}
\end{table}

\begin{table}[!htb] 
    \begin{minipage}[t]{0.49\textwidth} 
        \centering 
        \captionof{table}{Ablation study for alignment depth $D$ and regularization coefficient $\lambda$.}
        \label{tab:table3} 
        \tiny 
        \setlength{\tabcolsep}{2pt}
        \renewcommand{\arraystretch}{1.1} 
        \resizebox{0.99\textwidth}{!}{
            \begin{tabular}{l | >{\columncolor{gray!15}}c c c | c >{\columncolor{gray!15}}c c}
                \toprule
                \bfseries Metric & $D$=10   & $D$=20   & $D$=30     & $\lambda$=0.1 & $\lambda$=0.5 & $\lambda$=1  \\
                \midrule
                \bfseries FID ↓  & 27.61    & 28.41    & 31.84    & 27.81         & 27.61         & 27.76    \\
                \bfseries FIS ↑  & 0.638    & 0.591    & 0.503    & 0.634         & 0.638         & 0.627    \\
                \bfseries FVD ↓  & 180.52   & 230.95   & 423.51   & 189.72        & 180.52        & 191.78    \\
                \bottomrule
            \end{tabular}
        }
    \end{minipage}
    \hfill 
    \begin{minipage}[t]{0.49\textwidth} 
        \centering
        \captionof{table}{Ablation study for the design of CF-DiT block.} 
        \label{tab:table4}
        \tiny
        \setlength{\tabcolsep}{2pt}
        \renewcommand{\arraystretch}{1.1}
        \resizebox{0.99\textwidth}{!}{
        \begin{tabular}{l | *{5}{c}}
            \toprule
            \bfseries Settings                    & \bfseries FID↓  & \bfseries SSIM↑ & \bfseries PSNR↑ & \bfseries FIS↑ &  \bfseries FVD↓  \\
            \midrule
            DC Injection                & 30.68 & 0.682 & 29.09 & 0.544 & 372.80 \\
            CA Injection                & 32.30 & 0.675 & 28.85 & 0.502 & 453.73 \\ 
            \rowcolor{gray!15}
            DirectAnimator              & 27.61 & 0.752 & 29.53 & 0.638 & 180.52 \\
            \bottomrule
        \end{tabular}
        }
    \end{minipage}
\end{table}

\noindent \textbf{Same2X Training Strategy.}
To address the challenge of learning robust cross-identity human image animation, we introduce Same2X training strategy, a two-stage framework that first trains the model on same-identity data and then adapts it to cross-identity samples using a Same2X alignment loss.
As shown in Table \ref{tab:table2} (\textit{S8}), removing the Same2X strategy leads to consistent degradation across all metrics, including an increase in FID (+4.6), a noticeable drop in FIS (from 0.638 to 0.530), and a substantial increase in FVD (+110). These results underscore the contribution of Same2X training strategy during cross-ID training stage.
We further conduct ablation studies on two key hyperparameters: alignment depth ($D$) and regularization coefficient ($\lambda$). As shown in Table \ref{tab:table3}, increasing the alignment depth (e.g., $D$=20 and $D$=30) leads to significant performance drops. This aligns with REPA \citep{REPA}, indicating that shallow blocks capture geometric structure, while deeper blocks focus on high-frequency details. Applying the Same2X alignment loss to shallow blocks enhances regularization without hindering fine-grained learning in deeper blocks.
We also evaluate the impact of $\lambda$. As the results show, the model is relatively robust to different values of $\lambda$, and we adopt $\lambda = 0.5$ as the default setting.
Overall, these ablation studies validate the effectiveness and efficiency of the Same2X training strategy.

\noindent \textbf{CueFusion DiT Block.}
To validate the effectiveness of our proposed CueFusion DiT (CF-DiT) Block, we conduct an ablation study comparing it against two commonly used injection strategies:
(i) DC Injection, which \textbf{D}irectly \textbf{C}oncatenates the driving features with the noisy latent inputs to the DiT block;
(ii) CA Injection, which injects the driving features into the vision branch via \textbf{C}ross-\textbf{A}ttention.
As shown in Table \ref{tab:table4}, our CF-DiT Block consistently outperforms both DC and CA Injection across all metrics. This confirms the advantage of our design in effectively integrating driving cues into the denoising process.

\noindent \textbf{Pseudo Data Quality and Generator Choice.}
To further understand the impact of pseudo driving cues on Cross-ID training, we conduct an ablation study on pseudo data quality and generator choice.
As shown in Table~\ref{tab:table5}, using a large unfiltered pseudo set of 5,000 samples leads to noticeably worse performance compared to our filtered setup, indicating that simply increasing the amount of pseudo data without quality control can hurt both generation fidelity and temporal coherence.
When training with 500 pseudo cues from a single source, both \textit{MimicMotion-only} and \textit{StableAnimator-only} settings improve over the unfiltered baseline and achieve similar performance, suggesting that the model is not overly sensitive to the choice of pseudo generator and mainly relies on the shared motion and expression patterns.
Our \textit{Filtered pseudo set} achieves the best results across all metrics, demonstrating that our filtering strategy effectively controls pseudo bias while retaining the benefits of pseudo-driven training.

\begin{table}[tb]
    \centering
    \caption{
    Effect of pseudo data quality.
    }
    \label{tab:table5}
    \resizebox{0.75\textwidth}{!}
    {
    \large
    \begin{tabular}{l | *{5}{c}}
        \toprule
        \bfseries Settings                    & \bfseries FID↓  & \bfseries SSIM↑ & \bfseries PSNR↑ & \bfseries FIS↑ &  \bfseries FVD↓  \\
        \midrule
        Unfiltered pseudo set \textit{(5,000)}           & 29.93 & 0.712 & 29.17 & 0.584 & 218.30 \\
        MimicMotion-only pseudo cues \textit{(500)}         & 28.17 & 0.739 & 29.43 & 0.635 & 191.00 \\
        StableAnimator-only pseudo cues \textit{(500)}       & 28.22 & 0.745 & 29.40 & 0.630 & 187.40 \\

        \rowcolor{gray!15}
        Filtered pseudo set (Ours) \textit{(500)}     & 27.61 & 0.752 & 29.53 & 0.638 & 180.52 \\
        \bottomrule
    \end{tabular}
    }
\end{table}

\begin{table}[tb]
    \centering
    \caption{
    Inference time comparison on an H20 GPU at $512^2$ resolution (49 frames).
    }
    \label{tab:runtime}
    \resizebox{0.75\textwidth}{!}
    {
    \large
    \begin{tabular}{l | l l | c c c}
        \toprule
        \bfseries Method        & \bfseries Preproc      & \bfseries Backbone 
                                & \bfseries Preproc (ms) & \bfseries Gen (s)  \\
        \midrule
        AnimateAnyone           & DWpose & SD               & 9.6  & 102.59  \\
        StableAnimator          & DWpose & SVD              & 9.6  & 91.63  \\
        \rowcolor{gray!15}
        DirectAnimator (Ours)   & G-SAM + LPF & CogVideoX1.5           & 31  & 152.80 \\
        \bottomrule
    \end{tabular}
    }
\end{table}

\section{Conclusions}
We proposed DirectAnimator, a HIA framework that directly learns from raw driving videos without relying on pose estimators. By introducing a structured driving cue triplet and integrating it through the CueFusion DiT block, our method achieves high animation quality with improved robustness. To address the challenge of cross-ID training, we designed the Same2X strategy, which aligns internal representations between same-ID and cross-ID settings to guide learning. Extensive experiments confirm that DirectAnimator outperforms prior methods in visual quality and identity preservation. 
While DirectAnimator eliminates the need for pose estimators and curated annotations, limitations like pseudo cue quality, and explicit pose supervision remain. Future work aims to tackle these issues, paving the way for more intuitive and human-level animation learning.

\clearpage

\section*{Acknowledgements}
We thank the reviewers for the thoughtful discussion and feedback. 
This work was supported by the National Natural Science Foundation of China (U23B2030, Nos. 62311530100 and 62171251) and the Special Foundations for the Development of Strategic Emerging Industries of Shenzhen (No. KJZD20231023094700001).

\bibliography{iclr2026_conference}
\bibliographystyle{iclr2026_conference}

\clearpage

\appendix

\section{Appendix}
In this appendix, we first present the foundational concepts and diffusion-based architectures in Section~\ref{Fundamentals}. Section~\ref{DrivingCue} then provides an in-depth description of our Driving Cue representation, including the effect of low-pass filtering on pose cues, how spatial alignment is learned from pseudo cues, and the foreground-based spatial alignment used at inference time. In Section~\ref{Same2X}, we elaborate on the Same2X training strategy, including the generation and utilization of pseudo driving cues. Section~\ref{pose_metric} introduces our pose and expression evaluation metrics. Section~\ref{Visualization} presents additional visualizations, including more qualitative comparisons and further animation results, while Section~\ref{Failure} highlights representative failure cases. Finally, Section~\ref{Limitations} discusses the current limitations of our method and outlines promising directions for future work.

\subsection{Preliminaries}
\label{Fundamentals}
\paragraph{Diffusion Models.}  
Diffusion probabilistic models have emerged as a powerful class of generative models, framing data generation as the reversal of a gradual noise-adding process \citep{DDPM}. These models define a \textit{forward process} $q(\boldsymbol{x}_{1:T} | \boldsymbol{x}_0)$ that incrementally corrupts a clean data point $\boldsymbol{x}_0$ with Gaussian noise over $T$ discrete steps, followed by a \textit{reverse process} $p_{\theta}(\boldsymbol{x}_{0:T})$ that learns to denoise the noisy inputs back to the original data distribution.
In the \textit{forward diffusion process}, each step is defined as:
\begin{equation}
q(\boldsymbol{x}_t | \boldsymbol{x}_{t-1}) = \mathcal{N}(\boldsymbol{x}_t; \sqrt{1 - \beta_t} \boldsymbol{x}_{t-1}, \beta_t \boldsymbol{I}),
\end{equation}
where $\beta_t$ is the noise variance schedule at time $t$. This forms a Markov chain from $\boldsymbol{x}_0$ to $\boldsymbol{x}_T$, with $\boldsymbol{x}_T$ approaching an isotropic Gaussian distribution as $t \to T$.
For computational efficiency, we can directly sample a noisy version $\boldsymbol{x}_t$ at any time $t$ from $\boldsymbol{x}_0$ using the closed-form expression:
\begin{equation}
\boldsymbol{x}_t = \sqrt{\bar{\alpha}_t} \boldsymbol{x}_0 + \sqrt{1 - \bar{\alpha}_t} \epsilon, \quad \epsilon \sim \mathcal{N}(0, \boldsymbol{I}),
\end{equation}
where $\bar{\alpha}_t = \prod_{s=1}^t (1 - \beta_s)$ is the cumulative product of noise factors up to step $t$.
The \textit{reverse process} is parameterized by a neural network, often a time-aware U-Net, which estimates the mean and variance of the posterior $p_{\theta}(\boldsymbol{x}_{t-1} | \boldsymbol{x}_t)$:
\begin{equation}
p_{\theta}(\boldsymbol{x}_{t-1} | \boldsymbol{x}_t) = \mathcal{N}(\boldsymbol{x}_{t-1}; \mu_{\theta}(\boldsymbol{x}_t, t), \Sigma_{\theta}(\boldsymbol{x}_t, t)).
\end{equation}

In practice, the model is trained to predict the noise $\epsilon$ added at each step, using the simplified loss:
\begin{equation}
\mathcal{L}_{\text{simple}} = \mathbb{E}_{\boldsymbol{x}_0, t, \epsilon}\left[\left\| \epsilon - \epsilon_{\theta}(\boldsymbol{x}_t, t) \right\|^2 \right],
\end{equation}
where $\epsilon_{\theta}(\cdot)$ predicts the noise component in the sample $\boldsymbol{x}_t$ given the timestep $t$.
To enable \textit{conditional generation}, diffusion models can be extended with context vectors $c$ (e.g., text embeddings, segmentation masks), modifying the noise prediction to $\epsilon_{\theta}(\boldsymbol{x}_t, t, c)$. This enables controllable generation while maintaining sample diversity.
Despite their strong generation quality, diffusion models typically require \textit{hundreds of denoising steps} for high-fidelity samples, leading to slow inference. To address this, several acceleration methods have been proposed, such as DDIM \citep{DDIM} (non-Markovian deterministic sampling) and fast samplers based on knowledge distillation or noise schedule truncation.

\paragraph{Diffusion Transformers.}  
Diffusion Transformers (DiT) \citep{DiT} present a fully Transformer-based backbone for diffusion models, replacing the conventional convolutional U-Net architecture. Built upon the latent space framework of Stable Diffusion \citep{stablediffusion}, DiT processes image representations encoded by a fixed VAE encoder into low-dimensional feature maps. These latent tensors are segmented into non-overlapping patches and transformed into sequences of tokens via linear projection. Each token is enriched with temporal embeddings to represent the noise schedule step, along with optional conditioning inputs (e.g., class labels).
The model architecture comprises a stack of Transformer layers equipped with standard attention mechanisms, feed-forward networks, and normalization, all interleaved with residual paths. These components work together to model long-range dependencies and contextual relationships among latent patches. After traversing the Transformer pipeline, the output tokens are mapped back to the latent grid format, producing both noise and variance predictions. This dual-head output is then used to iteratively refine the denoising process across multiple diffusion steps. The design demonstrates strong scalability and performance.

\paragraph{CogVideoX Model.}  
Recent advances in diffusion-based generative models have significantly pushed the boundaries of video synthesis \citep{Cogvideox, Wan, Opensora}. Among these, CogVideoX \citep{Cogvideox} emerges as a scalable and high-performing diffusion Transformer architecture tailored for long-duration, text-conditioned video generation. Built upon the Diffusion Transformer (DiT) backbone~\citep{DiT}, CogVideoX integrates several critical innovations that address longstanding challenges in temporal coherence and cross-modal alignment. To efficiently encode the spatio-temporal redundancy in videos, CogVideoX adopts a 3D Causal Variational Autoencoder (VAE) which compresses video clips across both spatial and temporal axes, enabling tractable sequence lengths for Transformer modeling. Unlike prior works that employ frame-wise 2D VAEs, this 3D design notably improves motion continuity and reduces flickering artifacts. 
To bridge the semantic gap between text prompts and video content, CogVideoX introduces an Expert Transformer module, incorporating modality-specific adaptive LayerNorms that allow the model to conditionally modulate text and video representations. This design not only enhances alignment but also mitigates scale disparities in multimodal embeddings. Moreover, CogVideoX utilizes a 3D full attention mechanism to jointly capture spatial and temporal dependencies, in contrast to factorized attention used in prior approaches~\citep{Animatediff}, thus better preserving global scene dynamics.

Training CogVideoX involves a progressive paradigm including mixed-duration training, resolution scaling, and an innovative technique called explicit uniform sampling to stabilize loss convergence across diffusion timesteps. In addition, a comprehensive video-text data pipeline leveraging dense video captioning contributes to better semantic grounding. Empirical results demonstrate that CogVideoX outperforms previous state-of-the-art models across multiple automated metrics and human evaluations, particularly excelling in generating coherent, high-fidelity, and instruction-following videos. 

\subsection{More Details for Driving Cue}
\label{DrivingCue}
In the main paper, we introduced a novel Driving Cue representation that replaces conventional pose estimation with three complementary and semantically disentangled signals: \textit{Pose Cue}, \textit{Face Cue}, and \textit{Location Cue}. In this section, we provide detailed descriptions of the data processing pipelines used to generate each cue from raw driving videos. Additionally, we include more visual examples to illustrate the driving cue.

To facilitate learning of pose-related information, we first remove background from the driving video. Specifically, we employ the Grounded SAM model \citep{GroundedSAM} to segment out the foreground human subject. To further suppress redundant information, we apply a low-pass filter in the frequency domain to eliminate high-frequency image details. The resulting foreground image is used as the Pose Cue.
While most prior methods adopt 68 facial landmarks as the driving signal for expression transfer, such sparse representations often fail to capture complex facial expressions accurately, making it difficult to faithfully transfer expressions from the driving video to the reference image. Inspired by X-Dyna \citep{XDyna}, we instead propose to directly use the raw facial image as a more expressive input. However, in many driving frames, the face occupies only a small region of the image. To ensure the model can access high-resolution facial features, we utilize the InsightFace to detect the face and then enlarge and center it. The resulting image is used as the Face Cue.
The generated pose cue and face cue are not aligned to the spatial position of the reference identity, which may hinder the model's training process. To address this, we introduce an auxiliary signal for pose and face alignment. Specifically, we apply the alignment strategy from StableAnimator \citep{StableAnimator} to the pose and face masks in the driving video, aligning their spatial layout and scale with that of the reference identity. To prevent potential information leakage during training, we further apply a grid-based softening operation on the pose mask, blurring the mask boundaries while retaining the coarse silhouette. These aligned pose and face masks together form the Location Cue.

\begin{figure*}[h] \centering
    \includegraphics[width=0.95\textwidth]{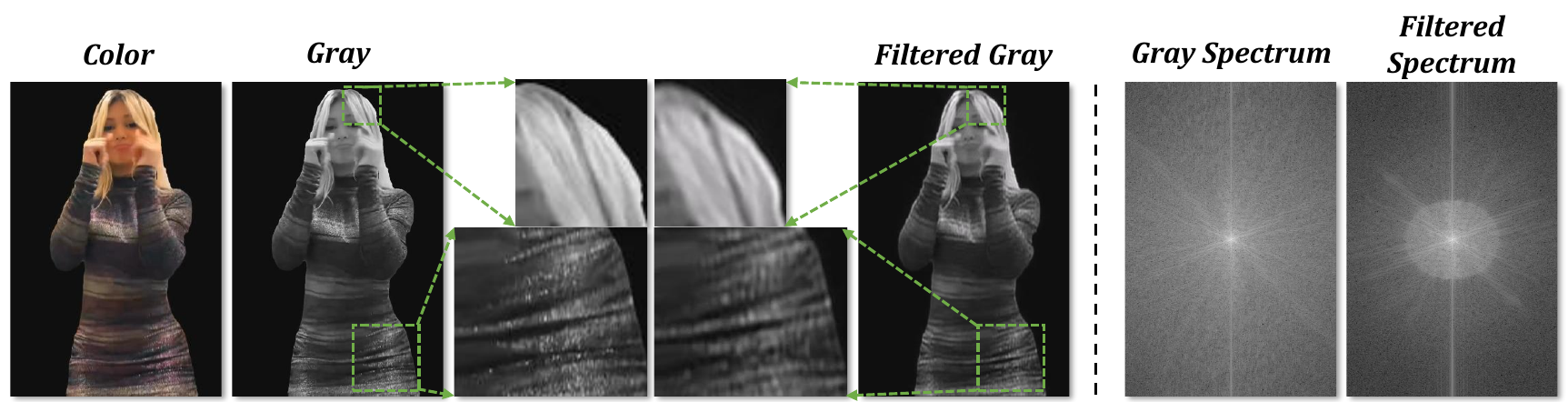}
    \caption{From left to right: original foreground color frame, grayscale frame with zoomed-in local patches, low-pass filtered grayscale frame with corresponding patches, and the log-magnitude Fourier spectra of the grayscale and filtered images. The spatial zooms reveal that fine hair and clothing textures are smoothed while the global body shape is preserved, and the spectra show that high-frequency components are strongly attenuated after low-pass filtering.}
    \label{fig:Fig_R4_1}
\end{figure*}

\noindent \textbf{Effect of Low-Pass Filtering on Pose Cues.}
To better illustrate the effect of the low-pass filter, Figure~\ref{fig:Fig_R4_1} visualizes a foreground frame before and after filtering in both the spatial and frequency domains. We first convert the segmented RGB frame to grayscale, which removes chromatic information but preserves most structural and textural details. Applying the low-pass filter in the frequency domain smooths fine textures on the hair and clothing while maintaining the global silhouette and coarse shading. This can be clearly seen from the zoomed-in patches, where high-frequency wrinkles and fabric patterns are noticeably attenuated in the filtered result. The log-magnitude Fourier spectra of the grayscale and filtered images further confirm this behavior: the filtered spectrum exhibits a strong concentration of energy around the low-frequency center and a substantial reduction of high-frequency components. These observations show that the pose cue indeed suppresses high-frequency appearance details while preserving the body structure required for motion control.

\noindent\textbf{Learning Spatial Alignment from Pseudo Cues.}
To learn robust spatial alignment, we do not directly use the raw outputs of StableAnimator \cite{StableAnimator} or MimicMotion \cite{Mimicmotion} as driving cues. 
First, we generate spatially aligned pseudo driving videos by selecting pseudo reference images whose body scale and location are close to the original driving videos, ensuring that these methods produce high-quality motion with only mild or no pose alignment. 
Then, after foreground segmentation, we apply additional scaling and translation to the segmented person to synthetically create spatially misaligned pseudo pose cues, while the Location Cue (pose and face masks) is extracted from the original driving video and remains aligned with the reference image. 
Training on such pairs of misaligned pseudo pose cues and aligned Location Cues explicitly teaches DirectAnimator to recover spatial alignment from the Location Cue, rather than inheriting any alignment behavior from StableAnimator or MimicMotion.

\noindent \textbf{Foreground-based Spatial Alignment at Inference Time.}
Inspired by the global similarity alignment in StableAnimator \cite{StableAnimator}, we align the driving sequence to the reference image directly at the foreground-mask level, without relying on skeleton keypoints.
Let $M^{\text{ref}} \in \{0,1\}^{H \times W}$ be the binary foreground mask of the reference image and $M_t \in \{0,1\}^{H \times W}$ be the foreground mask of the $t$-th driving frame.
For each mask $M$, we compute the tight person bounding box
\[
B = [x_{\min}, x_{\max}, y_{\min}, y_{\max}],
\]
and define its center and height as
\[
\mathbf{c}(B) = \bigl((x_{\min}+x_{\max})/2,\ (y_{\min}+y_{\max})/2\bigr)^\top,\quad
h(B) = y_{\max} - y_{\min}.
\]
Denote by $B^{\text{ref}}$ the bounding box of $M^{\text{ref}}$ and by $B_t$ the bounding box of $M_t$, with centers $\mathbf{c}^{\text{ref}} = \mathbf{c}(B^{\text{ref}})$ and $\mathbf{c}_t = \mathbf{c}(B_t)$, and heights $h^{\text{ref}} = h(B^{\text{ref}})$ and $h_t = h(B_t)$.
We first estimate a global isotropic scale $s$ by matching the person height over the whole driving sequence:
\begin{equation}
s_t = \frac{h^{\text{ref}}}{h_t},\quad
s = \operatorname{median}_t(s_t),
\label{eq:scale}
\end{equation}
and then align the centers by a single 2D translation vector $\mathbf{t} \in \mathbb{R}^2$:
\begin{equation}
\bar{\mathbf{c}} = \frac{1}{T}\sum_{t=1}^{T} \mathbf{c}_t,\quad
\mathbf{t} = \mathbf{c}^{\text{ref}} - s\,\bar{\mathbf{c}}.
\label{eq:translation}
\end{equation}
This yields a global similarity transform in the image plane,
\begin{equation}
\mathcal{T}(\mathbf{x}) = s\,\mathbf{x} + \mathbf{t}, \quad \mathbf{x} \in \mathbb{R}^2,
\label{eq:similarity-transform}
\end{equation}
which is shared by all driving frames.
We obtain the aligned foreground masks $\tilde{M}_t$ by warping each $M_t$ with the inverse transform $\mathcal{T}^{-1}$:
\begin{equation}
\tilde{M}_t(\mathbf{x}) = M_t\bigl(\mathcal{T}^{-1}(\mathbf{x})\bigr),
\quad \forall\,\mathbf{x} \in \{1,\dots,W\} \times \{1,\dots,H\}.
\label{eq:aligned-mask}
\end{equation}
In other words, the foreground person region in each driving frame is resized and shifted so that its location and scale are spatially aligned with the reference foreground region, while keeping a single, temporally stable similarity transform for the entire driving sequence.

\subsection{More Details for Same2X Training Strategy}
\label{Same2X}
The pseudo driving cues play a central role in the cross-ID training stage. They enable us to simulate diverse driving conditions for a given reference identity without relying on manually annotated or true cross-ID pairs. By leveraging motion and expression signals extracted from same-ID videos, these pseudo cues provide semantically meaningful supervision while avoiding identity leakage. 
In this section, we detail the process for generating pseudo driving cues used in the cross-ID training stage of our Same2X strategy. We generate two types of cues: pseudo pose cues and pseudo face cues.

To generate pseudo pose cues for the cross-ID training stage, we leverage StableAnimator~\citep{StableAnimator} and MimicMotion~\citep{Mimicmotion}, two of the most competitive human image animation methods to date. In addition to the data used for same-ID training, we collect an extra set of 1,000 images featuring diverse identities as the pseudo reference set. For each driving video sampled from the same-ID training set, we randomly select 0 to 3 images from the pseudo reference set as reference images. Using these pairs, we generate animated videos via the above animation models, followed by manual filtering to ensure quality. The filtered results serve as pseudo driving videos. We then apply foreground segmentation and low-pass filtering to obtain the corresponding pseudo pose cues.
Each pseudo driving cue thus preserves the same motion pattern as the original driving video but appears with a different identity and background. These samples can be interpreted as synthetic cross-ID driving signals for training, enabling the model to simulate cross-ID scenarios during training.


However, both StableAnimator and MimicMotion rely on skeleton maps to drive animation, which often results in inaccurate facial expressions compared to those in the original driving video. To address this limitation and obtain higher-fidelity facial dynamics, we adopt the face reenactment method Face-Adapter~\citep{FaceAdapter} to generate candidate facial animations. We then select the most visually consistent and expressive results for further processing.
Face reenactment aims to transfer facial expressions and head motion from a source video to a target identity, while preserving the target’s appearance and identity traits. In our case, pseudo reference images are used as target identities, and the driving videos serve as source inputs. We first crop the face regions from both inputs and apply Face-Adapter to transfer expressions from source to target. The resulting face regions are then centered and enlarged to construct the pseudo face cues, which provide complementary supervision for facial animation in the cross-ID training stage.

\begin{figure*}[h] \centering
    \includegraphics[width=0.99\textwidth]{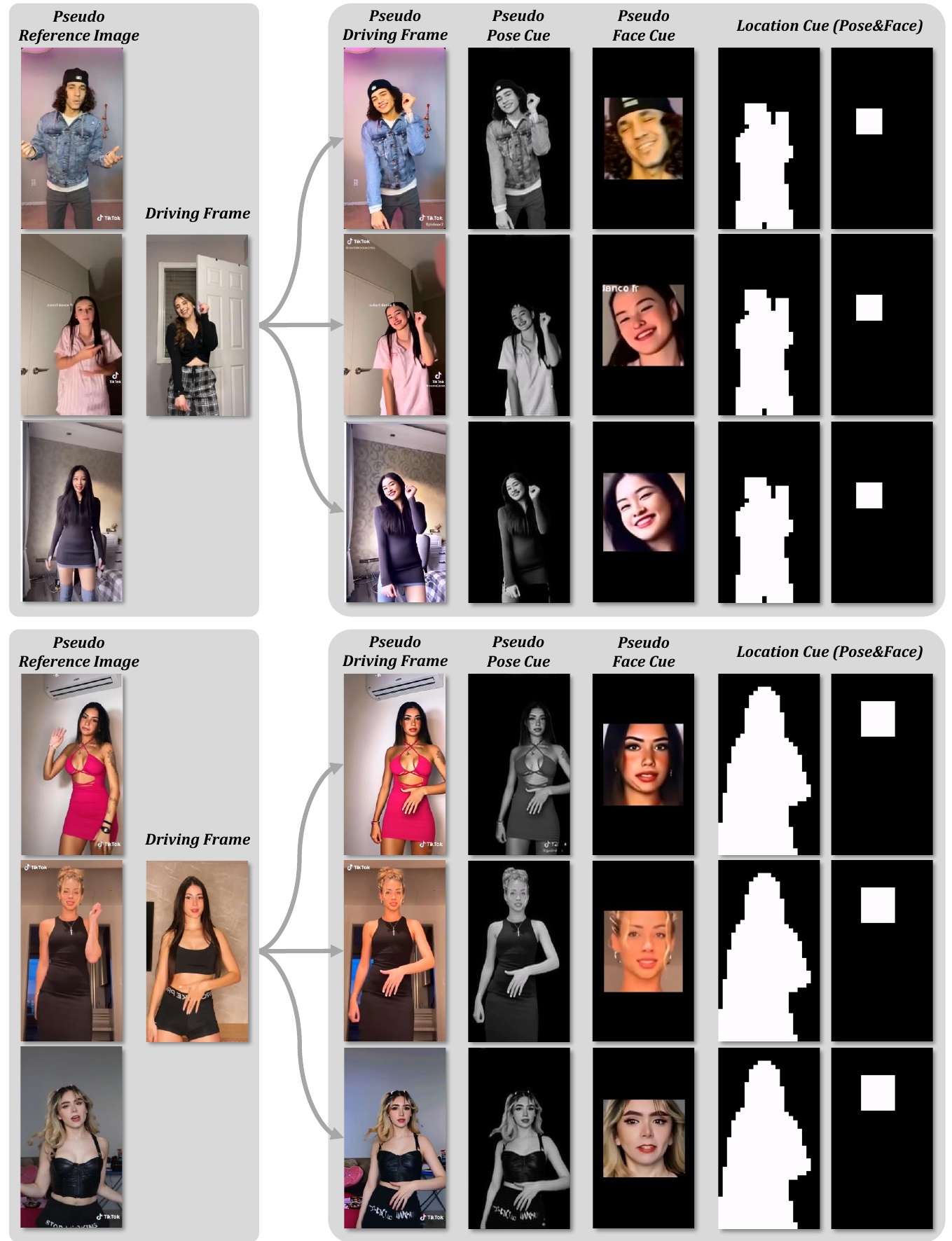}
    \caption{Examples of pseudo driving cues.}
    \label{fig:figure_s2}
\end{figure*}

For the Location Cue, we directly reuse the cues generated during the same-ID training stage. This is feasible because, in the cross-ID training stage, the reference images are sampled from the same-ID training data, and the driving signals (pseudo driving cues) are synthesized using driving videos from the same-ID stage. As a result, the pose and face alignment remain perfectly consistent, allowing us to reuse the same-ID location cues without additional processing.
Figure~\ref{fig:figure_s2} shows visual examples of the generated pseudo driving cues.

\begin{figure*}[h] \centering
    \includegraphics[width=0.99\textwidth]{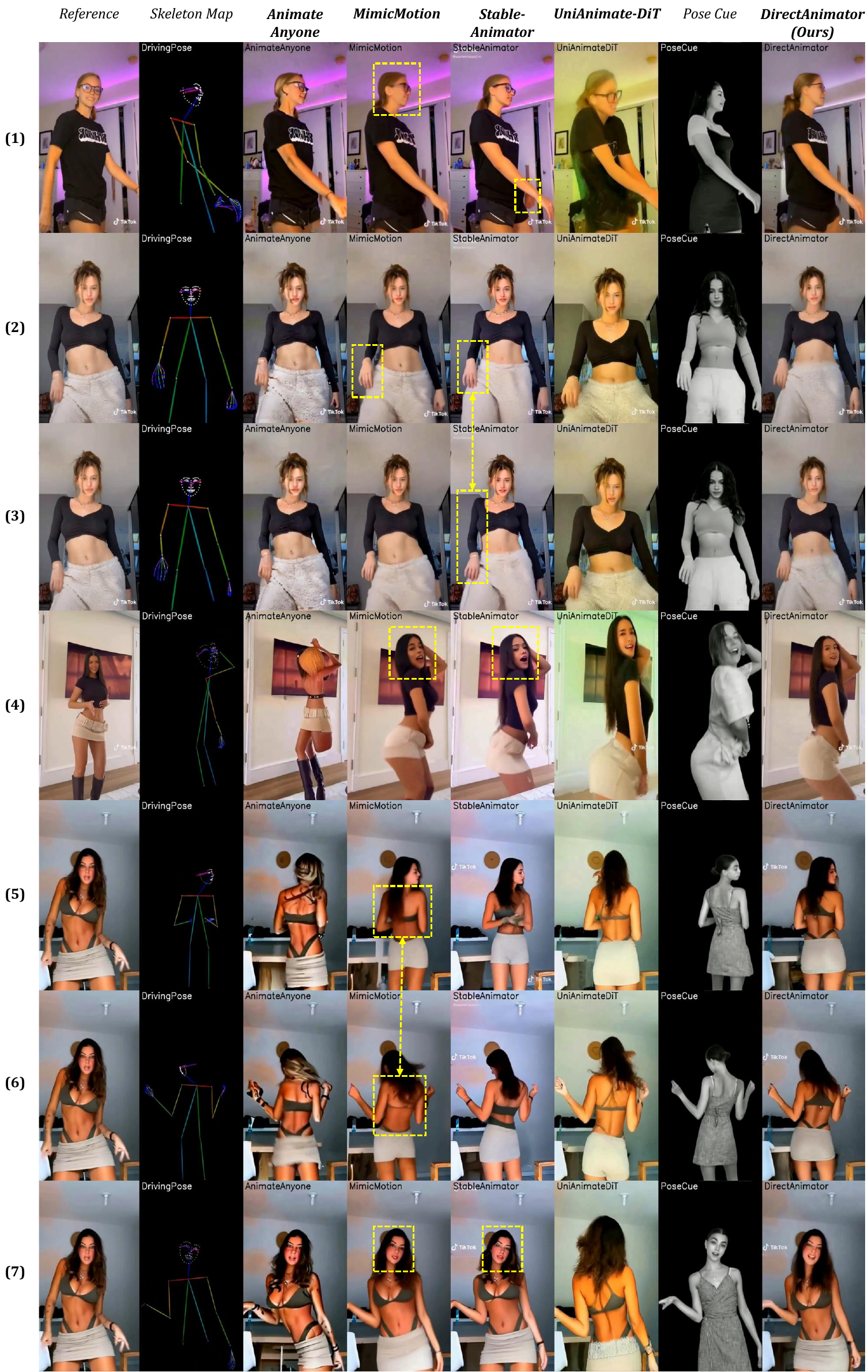}
    \caption{Qualitative comparisons with baseline methods, highlighting artifacts and showing the superiority of DirectAnimator in structural fidelity, temporal consistency, and expression transfer.}
    \label{fig:figure_s3}
\end{figure*}

\begin{figure*}[h] \centering
    \includegraphics[width=0.99\textwidth]{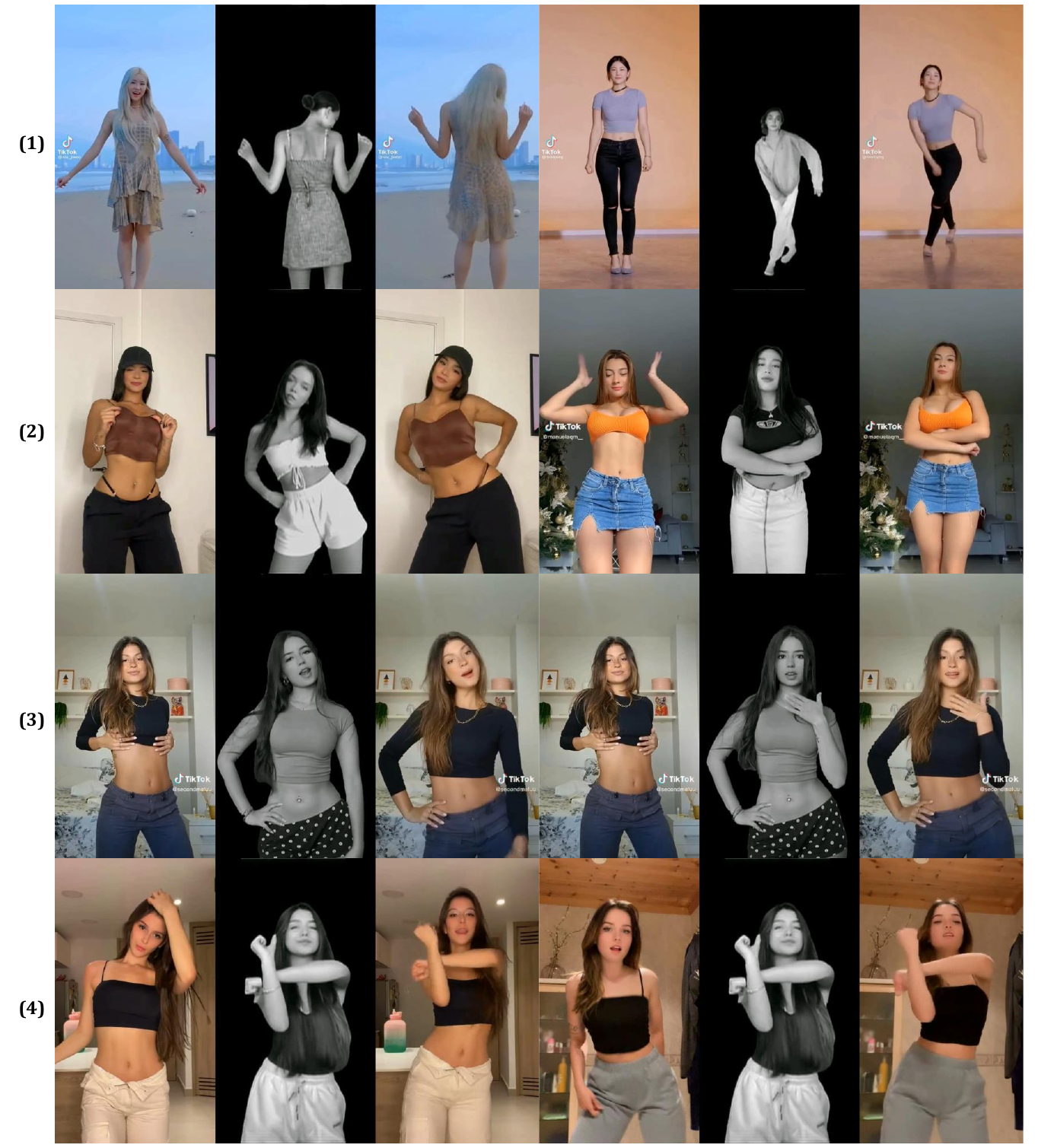}
    \caption{Animation results of DirectAnimator, demonstrating (1) pose alignment, (2) identity preservation, (3) expression transfer, and (4) complex motion modeling.}
    \label{fig:figure_s4}
\end{figure*}

\subsection{Pose and Expression Evaluation Metrics}
\label{pose_metric}
To comprehensively evaluate pose and facial expression transfer, we combine identity-based metrics with landmark-based geometric metrics.
\textit{Face Identity Similarity (FIS)} measures how well the identity of the reference image is preserved in the generated images. 
Concretely, we extract face embeddings from generated and reference images using ArcFace \citep{Arcface} and compute the average cosine similarity between corresponding pairs.
\textit{Face Temporal Similarity (FTS)} measures how temporally consistent the facial appearance remains within a generated video. 
We compute face embeddings for each frame using ArcFace and average the cosine similarity between embeddings of adjacent frames.
\textit{Pose Landmark Consistency (PLC)} measures how closely the generated body pose follows the driving pose. 
On the TikTok and Unseen test sets, we extract 2D body keypoints from both the driving and generated videos using DWpose, and compute the normalized distance between corresponding body landmarks.
\textit{Facial Landmark Consistency (FLC)} measures how accurately facial expressions and mouth shapes are transferred from the driving video. 
Similarly, we extract 2D facial keypoints from driving and generated frames and compute the normalized distance between corresponding facial landmarks.

Taken together, FIS/FTS and PLC/FLC offer a complementary view of model behavior: the former focuses on identity preservation and temporal smoothness, while the latter directly evaluates the geometric accuracy of pose and facial expression transfer. 
We report PLC and FLC in Table~\ref{tab:pose_face_metrics} and use them, alongside FIS/FTS and qualitative visualizations, to assess DirectAnimator on both TikTok and Unseen.

\subsection{More Visualization Results}
\label{Visualization}
\paragraph{More Qualitative Comparison.}
In Figure~\ref{fig:figure_s3}, we present more comprehensive comparison results, where the artifacts produced by baseline methods are highlighted with yellow bounding boxes for clarity. Specifically, in the first row, MimicMotion \cite{Mimicmotion} fails to capture fine facial details, while StableAnimator \cite{StableAnimator}, driven by an erroneous skeleton map, produces incorrect hand structures. In the second row, both MimicMotion and StableAnimator generate unnatural wrist structures, with wrist contours extending beyond the forearm. By contrast, our method faithfully reproduces anatomically correct wrist details. Moreover, StableAnimator exhibits temporal inconsistency in accessory generation: the bracelet is absent in the second row but reappears in the third row. In the fourth row, our approach achieves superior identity preservation, producing faces more consistent with the reference image. The fifth and sixth rows show that nearly all baselines struggle with body rotations, manifesting in structural errors and poor temporal coherence. We attribute this to the inherent limitations of skeleton map-based driving signals. In contrast, our method leverages the newly proposed driving cues, enabling more stable generation under such challenging scenarios. Finally, the seventh row demonstrates the stronger expression transfer capability of our model, with generated facial expressions that align more closely with those in the driving frames.

These comparisons clearly demonstrate that the proposed DirectAnimator outperforms existing methods across multiple dimensions, including structural fidelity, identity preservation, and temporal consistency. In particular, our method exhibits superior robustness and realism in scenarios involving complex body poses and expression changes, providing a more reliable solution for pose-free human image animation.

\paragraph{More Animation Results.}
In Figure~\ref{fig:figure_s4}, we present the animation results of DirectAnimator. The first row demonstrates the model’s capability in pose alignment: even when the driving frame and the reference image differ significantly in scale and position, DirectAnimator is able to generate high-quality animations. The second row highlights the strength of identity preservation, where the generated video frames maintain strong consistency with the reference image in the facial region and avoid distributional biases, faithfully retaining the phenotypic traits of the subject. The third row illustrates the model’s advantage in expression transfer, as the generated animations closely replicate the fine-grained facial expressions of the driving frame. Finally, the fourth row shows that DirectAnimator provides more robust modeling of complex motion structures: the generated video frames preserve the correct occlusion relationships between the two arms, whereas skeleton map-based approaches often confuse such relationships and degrade animation quality.
These results demonstrate that DirectAnimator consistently achieves superior performance across key dimensions, including pose alignment, identity preservation, expression transfer, and complex motion modeling.

\subsection{Failure Cases}
\label{Failure}
\begin{figure*}[h] \centering
    \includegraphics[width=0.99\textwidth]{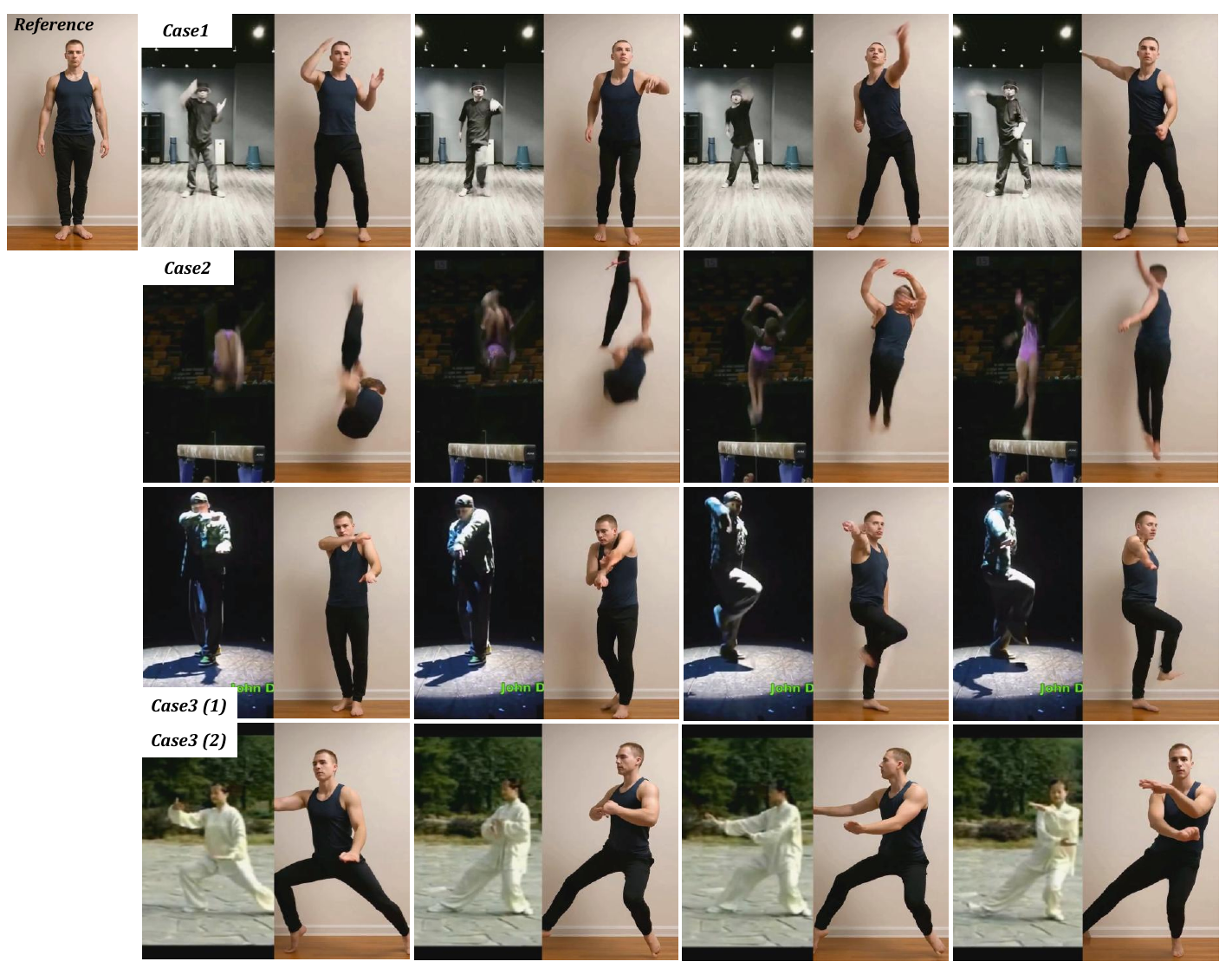}
    \caption{
    Failure cases of DirectAnimator. Case 1 shows loss of detail under mild motion blur, particularly in hands. Case 2 demonstrates structural distortion when motion blur obscures body configuration. Case 3 includes two subcases: inaccurate motion transfer under poor lighting (3(1)) and under low resolution (3(2)). 
    }
    \label{fig:failure_case}
\end{figure*}

We visualize and analyze several failure cases of DirectAnimator under challenging conditions. The corresponding videos are available on our project page. First, when the driving video contains mild motion blur, the generated frames often lose fine details in the affected regions. This issue is especially pronounced in hand articulation, as shown in Case 1 of Figure~\ref{fig:failure_case}. Second, when the driving video exhibits severe motion blur to the extent that human structure becomes ambiguous, such as during fast gymnastic movements, the generated animation may produce anatomically incorrect poses, as illustrated in Case 2. Third, low video quality also degrades performance. For example, poor lighting conditions as in Case 3(1) or low spatial resolution as in Case 3(2) make it difficult to accurately infer the subject's motion, resulting in noticeably lower animation quality.

\subsection{Limitations and Future Work}
\label{Limitations}
While DirectAnimator demonstrates strong performance across various benchmarks, several limitations remain.
First, since we use raw pixels from the driving video as the driving signal, motion blur in the input frames can affect the model’s ability to learn clean and precise features. As a result, the model may reproduce motion blur artifacts in the generated outputs, particularly under fast or low-light conditions.
Second, the visual fidelity is still bounded by the capacity of the underlying generative backbone. In some challenging cases, we occasionally observe imperfect identity preservation and noticeable artifacts, especially for fine-grained texture details.
Third, our method still relies on an explicit location cue to guide pose alignment during training. In contrast, humans naturally learn motion patterns through implicit alignment without requiring explicit location cues. Inspired by this observation, future work will explore removing the explicit pose alignment step. By improving training paradigms and incorporating larger and more diverse datasets, we aim to enable the model to learn implicit pose alignment in an end-to-end fashion, bringing HIA methods closer to human-level reasoning.
In future work, we also plan to enhance the pseudo cue generation process through more robust motion synthesis and automated quality filtering. Improving hand animation fidelity is another priority, potentially by integrating specialized hand pose estimators. Since DirectAnimator and Same2X are model-agnostic, another promising direction is to combine them with larger and higher-resolution video generation backbones, which we expect will further improve texture sharpness, clothing details, and identity preservation.

\section{Ethics Statement}
This work focuses on advancing human image animation through diffusion-based models. All training data were collected from publicly available sources, with user identities anonymized or obscured to ensure privacy protection. No personally identifiable information or sensitive attributes were used. Our method is designed for applications such as virtual avatars and creative content generation. Nevertheless, we acknowledge the potential for misuse in producing misleading or non-consensual content. To mitigate these risks, we encourage responsible use aligned with the ICLR Code of Ethics, and we explicitly discourage any application of our model that infringes upon personal rights or violates legal and ethical standards.

\section{Reproducibility Statement}
We have taken multiple steps to ensure reproducibility of our work. The main paper and appendix describe the model architecture, Driving Cue construction, and Same2X training strategy in detail. Implementation details, datasets, baselines, evaluation metrics, and ablation studies are also provided. In addition, we include the source code in the appendix files and provide an anonymous project page to facilitate independent verification of our results.

\section{LLM Usage Statement}
In this work, GPT-5 was utilized as a tool for language refinement. Specifically, the model was employed to improve the clarity, coherence, and readability of the manuscript. While GPT-5 played a important role in enhancing the presentation of the text, it was not involved in the ideation, experimental design, or data analysis processes. All content generated by the model was thoroughly reviewed and revised to ensure adherence to academic integrity and the originality of the work.

\end{document}